\documentclass[conference,letterpaper]{IEEEtran}

\usepackage[utf8]{inputenc} 
\usepackage[T1]{fontenc}
\usepackage{url}
\usepackage{ifthen}
\usepackage{cite}
\usepackage[cmex10]{amsmath} 

\usepackage{url}

\usepackage{booktabs} 
\usepackage{makecell}

\usepackage{soul}

\usepackage{amsfonts}       
\usepackage{nicefrac}       
\usepackage{microtype}      
\usepackage{xcolor}         
\usepackage[singlelinecheck]{caption}

\usepackage{graphicx}
\usepackage{times}
\usepackage{algorithm}
\usepackage{algorithmic}
\newtheorem{assumption}{Assumption}
\newtheorem{lemma}{Lemma}
\newtheorem{theorem}{Theorem}

\newtheorem{example}{Example}

\begin{document}
	\title{Rendering Wireless Environments Useful for Gradient Estimators: A Zero-Order Stochastic Federated Learning Method} 

	\author{%
		\IEEEauthorblockN{Elissa Mhanna and Mohamad Assaad}
		\IEEEauthorblockA{Laboratoire des Signaux et Systèmes\\
			Université Paris-Saclay, CNRS, CentraleSupélec\\
			91190 Gif-sur-Yvette, France \\
			Email: \{elissa.mhanna, mohamad.assaad\}@centralesupelec.fr}
	}

	\maketitle
		
	\begin{abstract}
	Cross-device federated learning (FL) is a growing machine learning setting whereby multiple edge devices collaborate to train a model without disclosing their raw data. With the great number of mobile devices participating in more FL applications via the wireless environment, the practical implementation of these applications will be hindered due to the limited uplink capacity of devices, causing critical bottlenecks. In this work, we propose a novel doubly communication-efficient zero-order (ZO) method with a one-point gradient estimator that replaces communicating long vectors with scalar values and that harnesses the nature of the wireless communication channel, overcoming the need to know the channel state coefficient. It is the first method that includes the wireless channel in the learning algorithm itself instead of wasting resources to analyze it and remove its impact. We then offer a thorough analysis of the proposed zero-order federated learning (ZOFL) framework and prove that our method converges \textit{almost surely}, which is a novel result in nonconvex ZO optimization. We further prove a convergence rate of $O(\frac{1}{\sqrt[3]{K}})$ in the nonconvex setting. We finally demonstrate the potential of our algorithm with experimental results.
	\end{abstract}
	\section{Introduction}
	ZO methods are a subfield of optimization that assume that first-order (FO) information or access to function gradients is unavailable. This arises in scenarios where exact gradient calculation may be expensive or unfeasible due to the inaccessibility of closed-form objectives. ZO optimization is based on estimating the gradient using function values queried at a certain number of points. The number of function queries depends on the assumptions of the problem. For example, in multi-point gradient estimates \cite{ref2, ref2-ref}, they construct the gradient by performing the difference of function values obtained at many random or predefined points. However, they assume that the stochastic setting stays the same during all these queries. By contrast, one-point estimates that use only one function value \cite{ref1, M-A}, principally obtained at a random point, 
	\begin{equation*}
		g= \frac{d}{\gamma}f(\theta + \gamma\Phi, S) \Phi,
	\end{equation*}
	assume that the settings are continuously changing during optimization. This is an important property as it resonates with many realistic applications, like when the optimization is performed in wireless environments or is based on previous simulation results. Recently, an appeal to ZO optimization is emerging in the machine-learning community, where optimizers are based on gradient methods. Examples include reinforcement learning \cite{RL2}, generating contrastive explanations for black-box classification models \cite{explainML}, and effecting adversarial perturbations on such models \cite{BBNN2}. 
	
	On the other hand, with the massive amounts of data generated or accessed by mobile devices, a growing research interest in both sectors of academia and industry is focused on FL \cite{tara, FL1}, as it is a practical solution for training models on such data without the need to log them to a server. A lot of effort has been invested in developing first-order \cite{FL1, FL2,FL3} and second-order methods  \cite{FLS, FLS2} to improve the efficacy of FL. These methods typically require access to the gradient or the Hessian of the local objective functions in their implementation to solve the optimization problem. However, using and exchanging such information raises many challenges, such as expensive communication and computation and privacy concerns \cite{FLchallenges}. 
	
	\begin{figure}[t]
		\centering
		\includegraphics[width=0.45\textwidth]{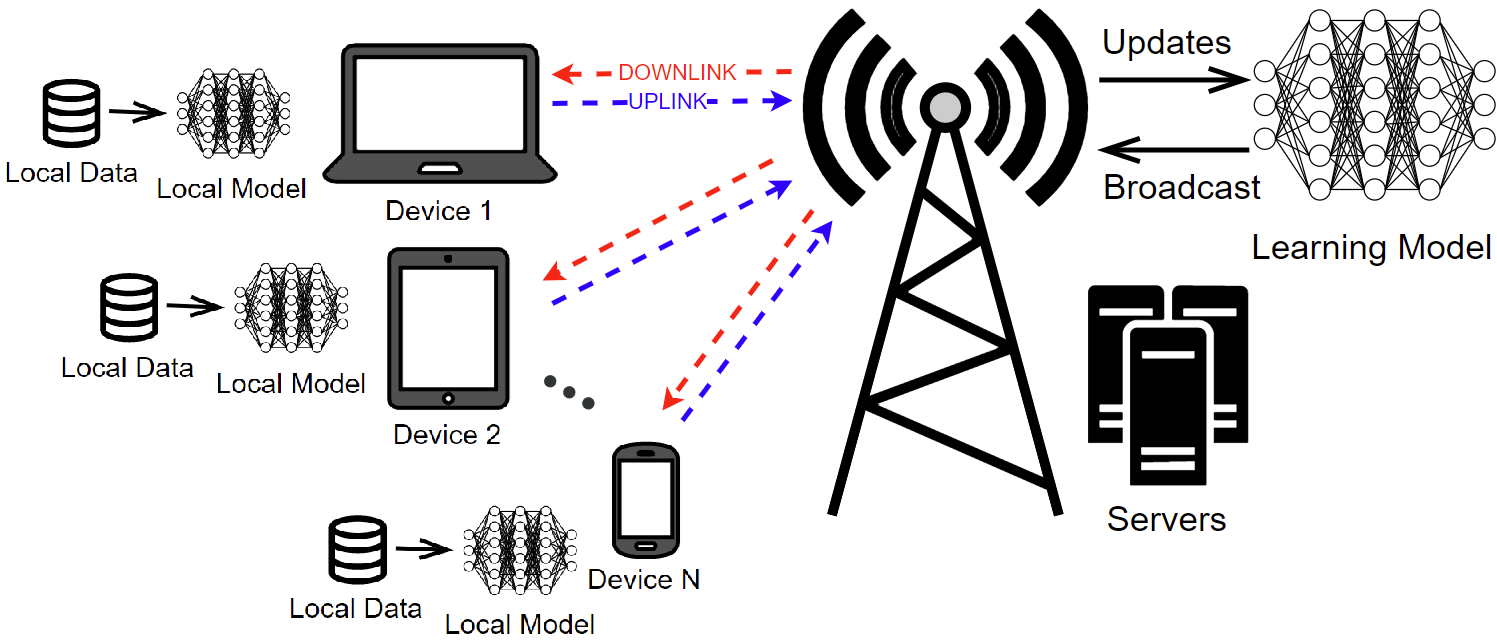}
		\caption{Federated learning over wireless networks.}
		\label{fig}
	\end{figure}
	There is more interest recently in learning over wireless environments \cite{com5, FL_wir1, FL_wir2, gunduz2}, with the increase of devices connected to servers through cellular networks. In this paper, we are interested in this scenario illustrated in Fig. \ref{fig}. Similarly to the aforementioned work, we are examining the case of analog communications between the devices and the server. However, it is a challenging problem as when the information is sent over the wireless channel, it becomes subject to a perturbation induced by the channel. This perturbation is not limited to additive noise, as noise is, in fact, due to thermal changes at the receiver. The channel acts as a filter for the transmitted signal \cite{tse},
	\begin{equation}\label{channel}
		\hat{x} = Hx+n
	\end{equation}
	where $x$ and $\hat{x}\in\mathbb{C}^d$ are the sent and received signals, respectively. $H\in\mathbb{C}^{d\times d}$ is the channel matrix, and $n\in\mathbb{C}^d$ is the additive noise, both of which are stochastic, constantly changing, and unknown. In general, all these entities are considered to have complex values. However, as we are interested in sending only real scalar values and we are not interested in decoding without error the sent information (but rather in perturbing the loss function), the phase shift introduced by the channel is considered as an additional perturbation in the channel model. Since our model is based on perturbing the loss function,  considering only the real part in the equation above is sufficient. We elaborate further on the channel modeling in Appendix \ref{channelmodel} for the interested reader. In FL, $x$ may denote the model or its gradients sent over the channel. To remove this impact, every channel element must be analyzed and removed to retrieve the sent information. This analysis is costly in computation and time resources. Thus, here our objective is to study FL in wireless environments without wasting resources. Further, we are interested in exploring the potential of ZO optimization to deal with some of the difficulties demonstrated by FL. 
	
	We thus consider an FL setting where a central server coordinates with $N$ edge devices to solve an optimization problem collaboratively. The data is private to every device and the exchanges between the server and the devices is restricted to the optimization parameters. To that end, let $\mathcal{N}=\{1,...,N\}$ be the set of devices and $\theta\in\mathbb{R}^d$ denote the global model. We define $F_i:\mathbb{R}^d\rightarrow\mathbb{R}$ as the loss function associated with the local data stored on device $i$, $\forall i\in\mathcal{N}$. The objective is to minimize the function $F:\mathbb{R}^d\rightarrow\mathbb{R}$ that is composed of the said devices' loss functions, such that 
	\begin{equation}\label{F_i}
		\min_{\theta\in\mathbb{R}^d} F(\theta):= \sum_{i=1}^{N} F_i(\theta)\;\;\text{with}\;\;	F_i(\theta)=\mathbb{E}_{S_i\sim D_i} f_i(\theta,S_i).
	\end{equation}
	
	$S_i$ is an i.i.d. ergodic stochastic process following a local distribution $D_i$. $S_i$ is used to model various stochastic perturbation, e.g. local data distribution among others. We further consider the case where the devices do not have access to their gradients for computational and communication restraints, and they must estimate this gradient by querying their model only once per update. They obtain a scalar value from this query (rather than a long vector as in the gradient), that they must send back to the server, which saves computation and communication resources since only a scalar is computed instead of a long vector of gradient/model as in the standard gradient approach. 
	\subsection{Motivation for Our Work}
	\textbf{Communication Bottleneck.} 
	In general, the main idea of FL is that the devices receive the model from the server, make use of their data to update the gradient, and then send back their gradients without ever disclosing their data. The server then updates the model using the collected and averaged gradients, and the process repeats. Since the gradients have the same dimension as the model, in every uplink step, there are $N d$ values that need to be uploaded, which forms the fundamental communication bottleneck in FL. To deal with this issue, some propose local multiple gradient descent steps to be done by the devices before sending their gradients back to the server to save communication resources \cite{localSGD}, or allow partial device participation at every iteration \cite{gunduz1}, or both \cite{FL1}. Others propose lossy compression of the gradient before uploading it to the server. For example, \cite{CQ3} suggests a stochastic unbiased quantization approach where gradients are approximated with a finite set of discrete values. \cite{CQ4} proposes the quantization of gradient differences between the current and previous iterations, allowing the update to incorporate new information, while \cite{S1} proposes the sparsification of this difference.
	
	\textbf{Channel Impact.} 
	In FL over wireless channels, there is a problem with channel knowledge. When the devices upload their gradient $g\in\mathbb{R}^d$ to the server, the server receives $Hg+n$ as shown in equation (\ref{channel}). In \cite{com5} and \cite{FedZO} and all references within, they assume that they can remove the impact of the channel. However, as the channel matrix $H$ coefficients follow a stochastic process and there are two unknown received entities, the channel $H$ and the gradient, the knowledge of the gradient requires estimating the channel coefficients at each  iteration of the FL. This requires computation resources, and more importantly, it requires resources to exchange  control/reference signals between the devices and the server at each time/iteration to estimate the channel coefficients $H$. Alternatively, our work offers a much simpler approach. We do not waste resources trying to analyze the channel. We use the channel in the implementation itself. It is part of the learning. We harness it to construct our gradient estimate without the need to remove its impact, saving both computation and communication resources. 
	
	\subsection{Contribution}	
	In this work, we propose a new communication-efficient algorithm in the nonconvex setting. This algorithm differs from the standard gradient method as it entails two reception-update steps instead of one, and it is not a simple extension of FO to ZO where the devices still have to upload their full model/gradient, as is the case in \cite{FedZO}. By limiting the exchange to scalar-valued updates, we counter the communication bottleneck, and we save up to a factor of $O(d)$ per communication round, in comparison to standard methods, in terms of total exchanges of variables between the devices and the server, saving a lot of execution time. We harness the nature of autocorrelated channels for truly "blind" reception of the data. We prove the convergence theoretically with a one-point estimate and provide experimental evidence. An important distinction worth noting is that standard ZO methods establish convergence by focusing on the expected convergence of the exact gradient. In contrast to prior research, our approach goes further in the proof. We demonstrate the convergence of the exact gradient itself almost surely, not solely its expected value. The key element in this proof is employing Doob's martingale inequality to constrain the stochastic error resulting from the estimated gradient. We further demonstrate a convergence rate of $O(\frac{1}{\sqrt[3]{K}})$ which outmatches centralized algorithms in the nonconvex ZO setting with one-point or noisy two-point estimates \cite{CZO, CZO2}.	While this rate matches our previous work \cite{icml}, the latter explores a ZO-distributed method aiming at achieving consensus. It entails a gradient tracking technique, which is a distinct algorithm that involves sharing between neighbors and updating two variables: the optimization variable and an auxiliary one. Thus, it has a worse communication cost as each user must send a long vector of $2d$ values, whereas here, only $2$ scalar values are sent per user. Furthermore, contrary to this work, the gradient tracking algorithm's performance is dependent on the network architecture, and the channel is not included in the algorithm. This simplifies the analysis regarding the gradient estimate and its bias as they're independent of the channel variables and the intertwined stochastic associations between them.
	
	\section{Algorithm}
	This section illustrates our proposed zero-order stochastic federated learning algorithm with a new gradient estimator (ZOFL). 
	\subsection{The 1P-ZOFL Algorithm}\label{1P-ZOFL}
	\begin{algorithm}[h]
		\caption{The 1P-ZOFL Algorithm}
		\label{alg:example}
		{\bfseries Input:} $\theta_0$, $\alpha_0$, $\gamma_{0}$, $\sigma_h$
		\begin{algorithmic}[1]
			\FOR{$k=0, 2, 4, ...$}
			\STATE The server receives $\sum_{j=1}^{N}\frac{h_{j,k}}{\sigma_h^2}+n_{j,k}$\\
			\STATE The server broadcasts $\theta_k'$ to all devices, \\ where $\theta_k'=\theta_k + \gamma_k\Phi_k\sum_{j=1}^{N}\big(\frac{h_{j,k}}{\sigma_h^2}+n_{j,k}\big)$\\
			\STATE The server receives\\ $\sum_{i=1}^{N}h_{i,k+1}\tilde{f}_i\big(\theta_k',S_{i,k+1}\big)+n_{i,k+1}$\\
			\STATE The server multiplies the received scalar sum by $\Phi_k$ to assemble $g_k$ given in (\ref{grdt_est})\\
			\STATE The server updates $\theta_{k+1} = \theta_k - \alpha_k g_k$
			\ENDFOR 
		\end{algorithmic}
	\end{algorithm}
	We consider an intermediary wireless environment between the server and each device $i$ for $i\in\mathcal{N}$ as shown in Fig. \ref{fig}. Wireless channels introduce a stochastic scaling on the sent signal as elaborated in equation (\ref{channel}). As we only send a scalar value over the channel at a time, our channel has only one scalar coefficient in addition to a scalar noise. Channel coefficients are usually autocorrelated from one timeslot to the next. Let $h_{i,k}$ denote the channel scaling affecting the sent signal from device $i$ to the server at timeslot $k$, independent from all other devices' channels. We assume $h_{i,k}$ to be a zero-mean random variable with standard deviation $\sigma_h$, $\forall i\in\mathcal{N}, \forall k\in\mathbb{N}^+$, and $n_{i,k}$ an additive noise on the transmitted signal. Assuming that the channel is time-correlated for two consecutive iterations $k$ and $k+1$, such that the autocovariance is $\mathbb{E}[h_{i,k}h_{i,k+1}]=K_{hh}$, $\forall i\in\mathcal{N}$, $\forall k\in\mathbb{N}^+$, we present our learning method in Algorithm \ref{alg:example}: 
	
	The devices must carry out two communication steps. In the first, every device sends the value $\frac{1}{\sigma_h^2}$ (or any other scalar value) to the server. According to equation (\ref{channel}), the server receives $\frac{h_{j,k}}{\sigma_h^2}+n_{j,k}$ from every device $j$. Hence, it receives the sum in step $2$. Afterward, the server uses the values received to adjust the model and broadcasts it to the devices. When device $i$ receives the model, it receives $h_{i,k+1}^{DL}[\theta_k + \gamma_k\Phi_k\sum_{j=1}^{N}(\frac{h_{j,k}}{\sigma_h^2}+n_{j,k})]+n_{i,k+1}^{DL}$, and to simplify notation, we let the stochastic vector $[h_{i,k+1}^{DL}, n_{i,k+1}^{DL}]$ be included within the big vector $S_{i,k+1}$ of stochastic perturbations. Device $i$ then queries this received model to obtain the stochastic loss $f_i$. Then the devices send $f$ to the server in the second communication step, and according to equation (\ref{channel}), the server receives the quantity indicated in step $4$. Finally, the server assembles the gradient estimate and is able to update $\theta$ according to step $7$. All transmissions are subject to channel scaling and additive noise. We designate them by $h$ and $n$ in the device-to-server transmission. In the server-to-devices one, we designate them by $S$. We let $\tilde{f}_i = \frac{f_i}{\sigma_h^2}$ be the normalized loss function and define $\alpha_k$ and $\gamma_k$ as two step-sizes and $\Phi_k\in\mathbb{R}^d$ as a perturbation vector generated by the server that has the same dimension as that of the model.
	
	We emphasize here that $g_k $ (in step $6$) is the gradient estimate in this case, and one can see that the impact of the channel is included in the gradient estimate and hence in the learning.
	The major advantage of this algorithm is that each device sends only two scalar values. This is stark improvement in communication efficiency over standard FL algorithms that require each device to send back the whole model or local gradient of dimension $d$. In effect, it is resource draining and can be unrealistic to assume it is possible.
	
	\subsection{The Estimated Gradient}\label{sub-grdt_est}
	We provide here analysis of our ZO gradient estimate. We propose the one-point estimate:
	\begin{equation}\label{grdt_est}
		g_k = \Phi_k \sum_{i=1}^{N}\Big[h_{i,k+1}\tilde{f}_i\Big(\theta_k', S_{i,k+1}\Big)+n_{i,k+1}\Big],
	\end{equation}
	where $\theta_k' = \theta_k + \gamma_k\Phi_k\sum_{j=1}^{N}\big(\frac{h_{j,k}}{\sigma_h^2}+n_{j,k}\big)$. The values $h_{i,k}$, $h_{i,k+1}$, and the noise remain unknown. This saves computation complexity and is very communication efficient as it transcends the need to send pilot signals to estimate the channel continuously. 
	
	We next consider the following assumptions on the additive noise, the perturbation vector, and the local loss functions. 
	\begin{assumption}\label{noise}
		$n_{i,k}$ is assumed to be a zero-mean uncorrelated noise with bounded variance, meaning $E(n_{i,k}) = 0$ and $E(n_{i,k}^2)=\sigma_n^2<\infty$, $\forall i\in\mathcal{N}$, $\forall k\in\mathbb{N}^+$. 
	\end{assumption}
	
	\begin{assumption}\label{perturbation}
		Let $\Phi_{k} = (\phi_{k}^1, \phi_{k}^2, \ldots, \phi_{k}^d)^T$.
		At each iteration $k$, the server generates its $\Phi_{k}$ vector independently from other iterations. In addition, the elements of $\Phi_{k}$ are assumed i.i.d with $\mathbb{E}(\phi_{k}^{d_1} \phi_{k}^{d_2}) =0$ for $d_1 \neq d_2$ and there exists $\beta_1 >0$ such that
		$\mathbb{E} (\phi_{k}^{d_j})^2 = \beta_1$, $\forall {d_j}$, $\forall k$.
		We further assume there exists a constant $\beta_2 >0$ where
		$\|\Phi_{k}\|\leq \beta_2$, $\forall k$.
	\end{assumption}
	\begin{example}\label{phi_eg} Generating $\Phi_k$ such that
		$\mathbb{P}(\phi_{k}^{d_j}=-\frac{1}{\sqrt{d}})=\mathbb{P}(\phi_{k}^{d_j}=\frac{1}{\sqrt{d}})=\frac12$,  $\forall {d_j}$, $\forall k$.
		 Then, $\beta_1=\frac{1}{d}$ and $\beta_2=1$.
	\end{example}
	
	\begin{assumption}\label{local_fcts}
		All loss functions $\theta\mapsto f_i(\theta,S_i)$ are Lipschitz continuous with Lipschitz constant $L_{S_i}$,
		$|f_i(\theta,S_i)-f_i(\theta',S_i)|\leq L_{S_i}\|\theta-\theta'\|$, $\forall i\in\mathcal{N}$.
		In addition, we assume there exists $C>0$ such that $|f_i (\theta, S_i)| \leq C < \infty, \forall i \in\mathcal{N}$.
	\end{assumption}
	
	Let $\mathcal{H}_k = \{\theta_0, S_0, \theta_1, S_1, ..., \theta_k, S_k\}$ denote the history sequence, then the following two lemmas characterize our gradient estimate. 
	\begin{lemma}\label{biased_estimators}
		Let Assumptions \ref{noise} and \ref{perturbation} be satisfied and define the scalar value $c_1=\beta_1\frac{K_{hh} }{\sigma_h^4}$, then the gradient estimator is biased w.r.t. the objective function's exact gradient $\nabla F(\theta)$. Concretely, 		
		$\mathbb{E}[g_k|\mathcal{H}_k] = c_1\gamma_k(\nabla F(\theta_k)+b_k)$,   
		$\forall k\in\mathbb{N}^+$, where $b_k$ is the bias term.
		
		Proof: Refer to Appendix \ref{app-grdt_est}.
		
		\begin{lemma}\label{norm}
			Let Assumptions \ref{noise}-\ref{local_fcts} hold. There exist a bounded constant $c_2 > 0$, such that $\mathbb{E}[\|g_k\|^2|\mathcal{H}_k] \leq c_2$.
			
			Proof: Refer to Appendix \ref{norm-sq}.
		\end{lemma}
	\end{lemma}
	
	\section{Convergence analysis}
	We begin by assuming that a global minimizer $\theta^*\in\mathbb{R}^{d}$ exists such that $\min_{\theta\in\mathbb{R}^d} F(\theta) = F(\theta^*)$ and $\nabla F(\theta^*)=0$. We then define $\delta_k = F(\theta_k)-F(\theta^{*})$ at every timeslot $k$.
	
	\begin{assumption}\label{objective_fct} 
		We assume the existence and the continuity of $\nabla F_i(\theta)$ and $\nabla^2 F_i(\theta)$, and that there exists a constant $\beta_3>0$ such that
		$\|\nabla^2 F_i(\theta)\|_2\leq\beta_3$,$\forall i\in\mathcal{N}$.

	\end{assumption}
	
	\begin{lemma}\label{smooth}
		By Assumption \ref{objective_fct}, we know that the objective function $\theta\longmapsto F(\theta)$ is $L$-smooth for some positive constant $L$,
		$\|\nabla F(\theta)-\nabla F(\theta')\|\leq L\|\theta-\theta'\|, \;\forall \theta,\theta'\in\mathbb{R}^d$.
	\end{lemma}
	
	\begin{lemma}\label{bias-norm-lemma}
		By Assumptions \ref{noise}, \ref{perturbation}, and \ref{objective_fct}, we can find a scalar value $c_3>0$ such that $\|b_k\| \leq c_3\gamma_k$.
		
		Proof: Refer to Appendix \ref{bias-norm}.
	\end{lemma}
	
	\subsection{1P-ZOFL convergence}
	In Lemma \ref{biased_estimators}, we see that in expectation, our estimator deviates from the gradient direction by the bias term. To provide that this term does not grow larger and preferably grows smaller as the algorithm evolves, we impose that $\gamma_k$ vanishes. Additionally, to ensure that the expected norm squared of the estimator, as shown in Lemma \ref{norm}, does not accumulate residual constant terms, we impose that the step size $\alpha_k$ vanishes. The series properties in the following assumption come from the recursive analysis of the algorithm. 
	
	\begin{assumption}\label{step_sizes_1}
		Both the step sizes $\alpha_k$ and $\gamma_k$ vanish to zero as $k\rightarrow\infty$ and the following series composed of them satisfy the convergence assumptions
		$\sum_{k=0}^{\infty}\alpha_k \gamma_k = \infty$, $\sum_{k=0}^{\infty}\alpha_k\gamma_k^3 <\infty$, and $\sum_{k=0}^{\infty}\alpha_k^2<\infty$.
	\end{assumption}
	\begin{example}\label{eg}
		Consider $\alpha_k = \alpha_0(1+k)^{-\upsilon_1}$ and $\gamma_k = \gamma_0 (1+k)^{-\upsilon_2}$ with $\upsilon_1, \upsilon_2>0$. Then, it is sufficient to find $\upsilon_1$ and $\upsilon_2$ such that $0<\upsilon_1+\upsilon_2\leq 1$, $\upsilon_1+3\upsilon_2>1$, and $\upsilon_1>0.5$.
	\end{example}
	We next define the stochastic error $e_k$ as the difference between the value of a single realization of $g_k$ and its conditional expectation given the history sequence, i.e., $e_k = g_k-\mathbb{E} [g_{k}|\mathcal{H}_k]$.
	
	The study of this noise and how it evolves is essential for the analysis of the algorithm as it gives access to the exact gradient when examining the algorithm's convergence behavior and permits us to prove that, in fact, the exact gradient converges to zero and not just the expectation of the exact gradient. This is a stronger convergence property, and it has not been done before in ZO nonconvex optimization to the best of our knowledge. The trick is to show that $e_k$ is a martingale difference sequence and to apply Doob's martingale inequality to derive the following lemma.
	
	\begin{lemma}\label{martingale}
		If all Assumptions \ref{noise}-\ref{step_sizes_1} hold, then for any constant $\nu>0$, we have
		$\mathbb{P}(\lim_{K\rightarrow\infty} \sup_{K'\geq K}\|\sum_{k=K}^{K'}\alpha_k e_k\|\geq\nu)=0$.
		
		Proof: Refer to Appendix \ref{martingale_proof}.
	\end{lemma}
	
	The smoothness inequality allows for the first main result, leading to the second in the following theorem.
	\begin{theorem}\label{th-ncvx}
		When Assumptions \ref{noise}-\ref{step_sizes_1} hold, then $\lim_{k\rightarrow\infty}\|\nabla F(\theta_k)\|=0$ almost surely, meaning that the algorithm converges.
		
		Proof: Refer to Appendix \ref{th-ncvx-proof}.
	\end{theorem}
	
	We next find an upper bound on the convergence rate of Algorithm \ref{alg:example}.
	\begin{theorem}\label{th-ncvx-rate}
		Consider in addition to the assumptions in Theorem \ref{th-ncvx}, that the step sizes are those of Example \ref{eg} with $\upsilon_1+\upsilon_2<1$. Then, for $\upsilon_1=\frac12+\frac{\epsilon}{2}$ and $\upsilon_2=\frac16+\frac{\epsilon}{2}$, we can write
		\begin{equation}\label{cv-rate}
			\begin{split}
				\frac{\sum_{k=0}^K\alpha_k\gamma_k\mathbb{E}\big[\|\nabla F(\theta_k)\|^2\big]}{\sum_{k=0}^K \alpha_k\gamma_k}	
				\leq O\left(\frac{1}{K^{\frac13-\epsilon}}\right).\\
			\end{split}
		\end{equation}
		
		Proof: Refer to Appendix \ref{th-ncvx-rate-proof}.
	\end{theorem}
	
	\section{Experimental results}
	
	For the experimental results, we test our algorithms in nonconvex binary image classification problems, and we compare them against the original FL algorithm FedAvg \cite{FL1} with exact gradient and one local update per round. However, \textit{we do not consider the effect of the channel or any noise/stochasticity for the FedAvg algorithm}. All experiments are done for $100$ devices and data batches of $10$ images per user per round. Every communication round in the graphs include all steps $2$ through $7$ for Algorithm \ref{alg:example}.
	
	We classify photos of the two digits ``$0$'' and ``$1$'' from the MNIST dataset using a nonconvex logistic regression model with a regularization parameter of $0.001$. All images are divided equally among the devices and are considered to be preprocessed by being compressed to have dimension $d=10$ using a lossy autoencoder. We run our code on $50$ simulations with different random model initializations testing the accuracy in every iteration against an independent test set. The graphs in Fig. \ref{iid} are averaged over all these simulations. For the non-IID data distribution, we first sort the images according to their labels and then divide them among the devices. $\Phi_k$ is generated according to Example \ref{phi_eg}. All channels are generated using the normal distribution with autocovariance $K_{hh}=\frac12$. The noise is Gaussian with $0$ mean and variance $\sigma_n^2=\frac14$.
	The step size for FedAvg is taken as $\eta=0.15$. For 1P-ZOFL, $\alpha_k = 0.5(1+k)^{-0.51}$ and $\gamma_k = 2.5(1+k)^{-0.18}$.
	Our algorithm performs consistently well with all the different random variations influencing every simulation in line with our theoretical results. Considering non-IID data distribution seems to be better impacted by our ZO algorithm slightly at the beginning without a major effect on the final result. 
	
	In Fig. \ref{iid2}, we test our algorithm with different noise variance $\sigma_n^2$ and notice that while the algorithm is slowed slightly at the beginning, it does not affect the convergence. When $\sigma_n^2=2.25$, we change $\alpha_0 = 0.1$ and $\gamma_0 = 0.8$. When $\sigma_n^2=10.0489$, $\alpha_0 = 0.07$ and $\gamma_0 = 0.3$. We provide 
	a discussion of our algorithm's performance in relation to the noise variance in Appendix \ref{experiments}.

	\begin{minipage}{0.45\textwidth}
		\includegraphics[width=\textwidth]{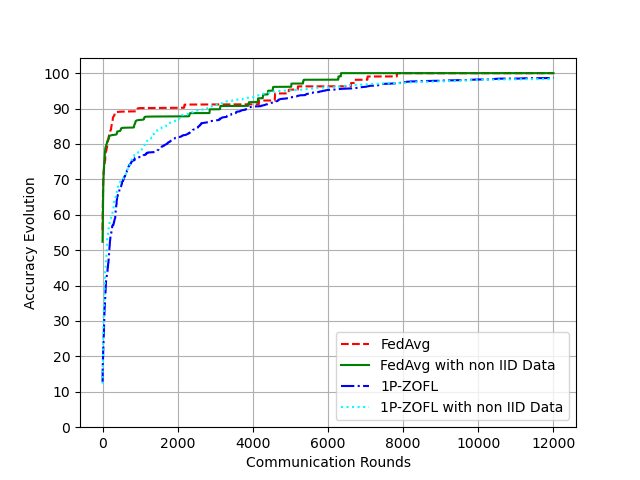}
		\captionof{figure}{Accuracy evolution of 1P-ZOFL vs. FedAvg for IID data and non-IID distribution.}
		\label{iid}
	\end{minipage}
	
	\begin{minipage}{0.45\textwidth}
		\includegraphics[width=\textwidth]{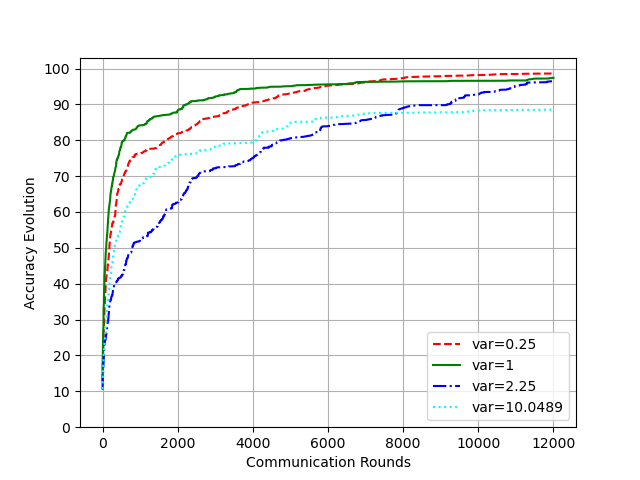}
		\captionof{figure}{Accuracy evolution of 1P-ZOFL for $\sigma_n^2 = \{0.25, 1, 2.25, 10.0489\}$.}
		\label{iid2}
	\end{minipage}

	 \section{Conclusion}
	 This work considers a learning problem over wireless channels and proposes a new zero-order federated learning method with a one-point gradient estimator. We limit the communication to scalar-valued feedback from the devices and incorporate the wireless channel in the learning algorithm. We provide theoretical and experimental evidence for convergence and find an upper bound on the convergence rate.
	 
	\bibliographystyle{IEEEtran}
	\bibliography{mybib}
	\appendices
	\section{On the channel model}\label{channelmodel}
	As described in Section 2.2 of Chapter 2 of \cite{tse} and considering a double-sideband suppressed carrier amplitude modulation (DSB-SC) instead of quadrature amplitude modulation (QAM):
	
	Having a baseband signal $x$, to send it over the channel, we modulate (multiply) it by $\sqrt{2}\cos2\pi f_c t$ where $f_c$ is the carrier frequency and $t$ is the time index.
	
	When sent over the channel, the transmitted signal $x$ undergoes perturbation and thus the received signal becomes:
	\begin{equation}\label{received}
		z = \sqrt{2}\sum_i a_i x \cos(2\pi f_c t+\varphi_i(t)) +w(t),
	\end{equation}
	where $a_i$ is the amplitude attenuation of path $i$ and $\varphi_i(t)=2\pi f_l t+\varphi_l$ is the phase shift incurred by Doppler frequency shift $f_l$ and/or any time delay $\varphi_l$. $w(t)$ is an additive noise.
	
	By developing the cosine term in $z$, we obtain
	\begin{equation}\label{received2}
		\begin{split}
			z = &x\sqrt{2}\underbrace{\sum_i a_i \cos(\varphi_i(t))}_{\text{in-phase component, }I(t)}\cos(2\pi f_c t)\\&-x\sqrt{2}\underbrace{\sum_i a_i\sin(\varphi_i(t))}_{\text{quadrature component, }Q(t)} \sin(2\pi f_c t)+w(t),
		\end{split}
	\end{equation}
	From Section $2.4.2$ of \cite{tse}: According to the central limit theorem, if there is a large number of channel paths, the in-phase and quadrature components of the received signal, which are not correlated with each other, will exhibit distributions that resemble the normal (Gaussian) distribution. Specifically, each component will have an average value of zero and a variance of $\Sigma/2$, which is equivalent to $\sigma^2$. The magnitude of the perturbation $\sqrt{I(t)^2+Q(t)^2}$ thus becomes Rayleigh distributed. This is the Rayleigh fading model. In addition, when the line-of-sight path is large and has a known magnitude, the probabilistic model becomes a Rician fading.
	
	Furthermore, as $I(t)$ and $Q(t)$ are orthogonal, an equivalent complex channel model $\hat{h}(t)=I(t)+j Q(t)=a(t)e^{-j\varphi(t)}$ can be derived. Since the carrier frequency $f_c$ is not involved in $\hat{h}(t)$, this representation is valid at baseband level. Thus, the complex channel model is usually used to represent the received signal $\hat{h}(t)x+n(t)$ at baseband with $\hat{h}(t)$ a complex entity.
	
	Continuing from (\ref{received2}), to demodulate $z$ and obtain the baseband received signal $y$, $z$ is first multiplied by $\sqrt{2}\cos2\pi f_c t$ then the result goes through a low pass filter.
	
	\begin{equation*}
		\begin{split}
			&z \sqrt{2}\cos(2\pi f_c t)\\ = &2x\sum_i a_i \cos(\varphi_i(t))\cos^2(2\pi f_c t)\\&-2x\sum_i a_i\sin(\varphi_i(t)) \sin(2\pi f_c t)\cos(2\pi f_c t)\\&+\sqrt{2}w(t)\cos(2\pi f_c t)\\
			= &x\sum_i a_i \cos(\varphi_i(t))(1+\cos(4\pi f_c t))\\&-x\sum_i a_i\sin(\varphi_i(t)) \sin(4\pi f_c t)+\sqrt{2}w(t)\cos(2\pi f_c t) \\
		\end{split}
	\end{equation*}
	After the low pass filter, we obtain the received baseband signal
	\begin{equation}\label{received3}
		\begin{split}
			y = &x\sum_i a_i \cos(\varphi_i(t))+n(t)\\
			= &\mathfrak{R}[\hat{h}(t)]x+n(t)\\
			= &h(t)x+n(t)
		\end{split}
	\end{equation}
	where $n(t)$ is the baseband equivalent noise with a zero-mean Gaussian distribution and IID components (Section $2.2.4 $ of \cite{tse}) and $\mathfrak{R}[\hat{h}(t)] = I(t)$ is the real part of the channel. 
	
	As we are interested to send real values over the wireless channel in this paper, one can easily see how equation (\ref{received3}) is valid to use with a real channel $h = \mathfrak{R}[\hat{h}]$ following a Gaussian distribution with zero mean and variance equal to $\sigma^2$.
	
	\section{On the Estimated Gradient}
	\subsection{Proof of Lemma \ref{biased_estimators}: Biased Estimator}\label{app-grdt_est}
	Let $g_k$ have the form in (\ref{grdt_est}), then the conditional expectation given $\mathcal{H}_k$ can be written as
	\begin{equation*}
		\begin{split}
			&\mathbb{E}[g_k|\mathcal{H}_k] \\
			= &\mathbb{E}\Big[\Phi_k \sum_{i=1}^{N}\Big(h_{i,k+1}\tilde{f}_i\big(\theta_k', S_{i,k+1}\big)+n_{i,k+1}\Big)\Big|\mathcal{H}_k\Big]\\
			\overset{(a)}{=} &\mathbb{E}\Big[\Phi_k \sum_{i=1}^{N}h_{i,k+1}\tilde{F}_i\big(\theta_k + \gamma_k\Phi_k\sum_{j=1}^{N}\big(\frac{h_{j,k}}{\sigma_h^2}+n_{j,k}\big)\big)\Big|\mathcal{H}_k\Big]\\
			\overset{(b)}{=} &\mathbb{E}\Big[\Phi_k\Big( \sum_{i=1}^{N}h_{i,k+1}\tilde{F}_i(\theta_k)\\
			&+\gamma_k\sum_{i=1}^{N}h_{i,k+1}\sum_{j=1}^{N}\Big(\frac{h_{j,k}}{\sigma_h^2}+n_{j,k}\Big)\Phi_k^T \nabla\tilde{F}_i(\theta_k)\\
			&+\gamma_k^2\sum_{i=1}^{N}h_{i,k+1}\Big(\sum_{j=1}^{N}\frac{h_{j,k}}{\sigma_h^2}+n_{j,k}\Big)^2 \Phi_k^T \nabla^2 \tilde{F}_i(\breve{\theta}_k)\Phi_k\Big)\Big|\mathcal{H}_k\Big]\\
			\overset{(c)}{=} &\mathbb{E}\Big[\Phi_k \bigg(\frac{\gamma_k}{\sigma_h^2}\sum_{i=1}^{N}h_{i,k+1}h_{i,k}\Phi_k^T\nabla\tilde{F}_i(\theta_k)\\
			&+\gamma_k^2\sum_{i=1}^{N}h_{i,k+1}\Big(\sum_{j=1}^{N}\frac{h_{j,k}}{\sigma_h^2}+n_{j,k}\Big)^2 \Phi_k^T \nabla^2 \tilde{F}_i(\breve{\theta}_k)\Phi_k\Big)\Big|\mathcal{H}_k\Big]\\
		\end{split}
	\end{equation*}
	\begin{equation}\label{bias_1P}
		\begin{split}
			= &\frac{\gamma_k}{\sigma_h^2} 	\sum_{i=1}^{N}\mathbb{E}\Big[h_{i,k+1}h_{i,k}\Big|\mathcal{H}_k\Big]\mathbb{E}\Big[\Phi_k \Phi_k^T\Big|\mathcal{H}_k\Big]\nabla\tilde{F}_i(\theta_k)+\\
			&\gamma_k^2\sum_{i=1}^{N}\mathbb{E}\Big[h_{i,k+1}\Big(\sum_{j=1}^{N}\frac{h_{j,k}}{\sigma_h^2}+n_{j,k}\Big)^2 \Phi_k\Phi_k^T \nabla^2 \tilde{F}_i(\breve{\theta}_k)\Phi_k\Big|\mathcal{H}_k\Big]\\
			\overset{(d)}{=} &\gamma_k\beta_1\frac{K_{hh} }{\sigma_h^4}  \sum_{i=1}^{N} \nabla F_i(\theta_k)+\\ &\gamma_k^2\sum_{i=1}^{N}\mathbb{E}\Big[h_{i,k+1}\Big(\sum_{j=1}^{N}\frac{h_{j,k}}{\sigma_h^2}+n_{j,k}\Big)^2 \Phi_k\Phi_k^T \nabla^2 \tilde{F}_i(\breve{\theta}_k)\Phi_k\Big|\mathcal{H}_k\Big]\\
			\overset{(e)}{=} &c_1\gamma_k(\nabla F(\theta_k)+b_k),
		\end{split}
	\end{equation}
	where $(a)$ is by the definition in (\ref{F_i}) and due to Assumption \ref{noise}, $(b)$ is by Taylor expansion and mean-value theorem and considering $\breve{\theta}_k$ between $\theta_k$ and $\theta_k + \gamma_k\Phi_k\sum_{j=1}^{N}\big(\frac{h_{j,k}}{\sigma_h^2}+n_{j,k}\big)$. $(c)$ is since $\mathbb{E}[h_{i,k+1}]=0$ for the first element and $\mathbb{E}[h_{i,k+1}h_{j,k}]=0$ when $i\neq j$ and the independence of noise for the second element. $(d)$ is due to Assumption \ref{perturbation}. In $(e)$, we let $c_1=\beta_1\frac{K_{hh} }{\sigma_h^4}$ and the bias 
	$b_k=\gamma_k \frac{\sigma_h^2}{\beta_1 K_{hh}}\sum_{i=1}^{N}\mathbb{E}\Big[h_{i,k+1}\Big(\sum_{j=1}^{N}\frac{h_{j,k}}{\sigma_h^2}+n_{j,k}\Big)^2 \Phi_k\Phi_k^T \nabla^2 \tilde{F}_i(\breve{\theta}_k)\Phi_k\Big|\mathcal{H}_k\Big]$.
	\subsection{Proof of Lemma \ref{norm}: Expected Norm Squared of the Estimated Gradient}\label{norm-sq}
	\begin{equation}\label{grdt_norm_1}
	\begin{split}
			&\mathbb{E}[\|g_k\|^2|\mathcal{H}_k]\\
			= &\mathbb{E}\Big[\|\Phi_k \|^2\Big(\sum_{i=1}^{N}h_{i,k+1}\tilde{f}_i\big(\theta_k', S_{i,k+1}\big)+n_{i,k+1}\Big)^2\Big|\mathcal{H}_k\Big]\\
			\overset{(a)}{\leq} &\beta_2^2 \mathbb{E}\Big[\Big(\sum_{i=1}^{N}h_{i,k+1}\tilde{f}_i\big(\theta_k', S_{i,k+1}\big)+n_{i,k+1}\Big)^2\Big|\mathcal{H}_k\Big]\\
			= &\beta_2^2 \mathbb{E}\Big[\Big(\sum_{i=1}^{N}h_{i,k+1}\tilde{f}_i\big(\theta_k', S_{i,k+1}\big)\Big)^2+\Big(\sum_{i=1}^{N}n_{i,k+1}\Big)^2\\&\hspace{1cm}+2\Big(\sum_{i=1}^{N}h_{i,k+1}\tilde{f}_i\big(\theta_k', S_{i,k+1}\big)\Big)\Big(\sum_{i=1}^{N}n_{i,k+1}\Big)\Big|\mathcal{H}_k\Big]\\
			\overset{(b)}{=} &\beta_2^2 \mathbb{E}\Big[\Big(\sum_{i=1}^{N}h_{i,k+1}\tilde{f}_i\big(\theta_k', S_{i,k+1}\big)\Big)^2+\Big(\sum_{i=1}^{N}n_{i,k+1}\Big)^2\Big|\mathcal{H}_k\Big]\\	
			\overset{(c)}{=} &\beta_2^2 \mathbb{E}\Big[N\sum_{i=1}^{N}\Big(h_{i,k+1}^2\frac{C^2}{\sigma_h^4}\Big)+\sum_{i=1}^{N}n_{i,k+1}^2\\&\hspace{1cm}+2\sum_{j<i}n_{i,k+1}n_{j,k+1}\Big|\mathcal{H}_k\Big]\\	
			\overset{(d)}{=} &\beta_2^2 \sum_{i=1}^{N}\mathbb{E}\Big[N h_{i,k+1}^2\frac{C^2}{\sigma_h^4}+n_{i,k+1}^2\Big|\mathcal{H}_k\Big]\\	
			= &\beta_2^2 N \Big(\frac{N C^2}{\sigma_h^2}+\sigma_n^2\Big)\\
			:= &c_2,
		\end{split}
	\end{equation}
	where $(a)$ is by Assumption \ref{perturbation}, $(b)$ is due to the due to the independence of $n_{i,k+1}$ and its zero-mean, $(c)$ is by Cauchy-Schwartz $(\sum_{j=1}^{N}1\cdot a_i)^2 \leq (\sum_{j=1}^{N}a_i^2)\cdot (\sum_{j=1}^{N}1^2 )  = N (\sum_{j=1}^{N}a_i^2) $, for real $a_i$, and Assumption \ref{local_fcts}. $(d)$ is due to the uncorrelatedness of the noise.
	
	\subsection{Proof of Lemma \ref{bias-norm-lemma}: Norm of the Bias} \label{bias-norm}
	Considering the form of the bias in (\ref{bias_1P}), by Assumptions \ref{noise}, \ref{perturbation} and \ref{objective_fct},
	\begin{equation*}
		\begin{split}
			\|b_k\|\overset{(a)}{\leq} &\gamma_k \frac{\sigma_h^2}{\beta_1 K_{hh}} \sum_{i=1}^{N}\mathbb{E}\Big[2N \Big|h_{i,k+1}\sum_{j=1}^{N}\Big(\frac{h_{j,k}^2}{\sigma_h^4}+n_{j,k}^2\Big)\Big|\times\\&\hspace{2.85cm}\|\Phi_k\| \|\Phi_k^T\| \|\nabla^2 \tilde{F}_i(\breve{\theta}_k)\| \|\Phi_k\| \Big|\mathcal{H}_k\Big]\\
			\leq &2 \gamma_k \frac{N\beta_3\beta_2^3\sigma_h^2}{\beta_1 K_{hh}} \sum_{i=1}^{N}\mathbb{E}\Big[\Big|h_{i,k+1}\sum_{j=1}^{N}\Big(\frac{h_{j,k}^2}{\sigma_h^4}+n_{j,k}^2\Big)\Big|\Big|\mathcal{H}_k\Big]\\
	\end{split}
	\end{equation*}
	\begin{equation}\label{bias_norm}
		\begin{split}
			\overset{(b)}{\leq} &2 \gamma_k \frac{N\beta_3\beta_2^3\sigma_h^2}{\beta_1 K_{hh}} \sum_{i=1}^{N}\Big[\sigma_h\sqrt{\frac{2}{\pi}}\Big(2K_{hh}+\sqrt{\sigma_h^4-K_{hh}^2}\Big)\\&\hspace{2.8cm}+(N-1)\sigma_h^3\sqrt{\frac{2}{\pi}}+N\sqrt{\frac{2}{\pi}}\sigma_h\sigma_n^2\Big]\\
			=&2 \gamma_k \frac{N^2\beta_3\beta_2^3\sigma_h^3}{\beta_1 K_{hh}} \sqrt{\frac{2}{\pi}}\times\\&\hspace{0.8cm}\Big[\Big(2K_{hh}+\sqrt{\sigma_h^4-K_{hh}^2}\Big)+(N-1)\sigma_h^2+N\sigma_n^2\Big]\\
			:= &c_3\gamma_k,
		\end{split}
	\end{equation}
	where $(a)$ is due to Jensen's inequality and Cauchy-Schwartz. $(b)$ is by using the half-normal distribution for normal random variables in absolute value:	
	Let $X$ and $Y$ be two variables representing time-correlated channel realizations at times $k$ and $k'$ respectively. Assume they follow the $\mathcal{N}(0,\sigma)$ distribution and they have a correlation coefficient $\varrho$. Then, we can write $Y= \varrho X + \sqrt{1-\varrho^2} Z$, where $Z$ is independent of X and following the same distribution $\mathcal{N}(0,\sigma)$.
	Then,
	$\mathbb{E}[|YX^2|] 
	\leq \mathbb{E}[\varrho |X^3| + \sqrt{1-\varrho^2} |Z X^2|] =2\varrho\sqrt{\frac{2}{\pi}}\sigma^3 +  \sqrt{1-\varrho^2}\sqrt{\frac{2}{\pi}}\sigma\times \sigma^2= (2\varrho+ \sqrt{1-\varrho^2})\sqrt{\frac{2}{\pi}}\sigma^3.$
	If we substitute $\sigma=\sigma_h$ and $\varrho=\frac{K_{hh}}{\sigma_h^2}$, we obtain the previous inequality $(b)$.
	\section{1P-ZOFL algorithm convergence}
	\subsection{Stochastic Noise}\label{martingale_proof} 
	To prove Lemma \ref{martingale}, we begin by demonstrating that the sequence $\{\sum_{k=K}^{K'} \alpha_k e_k\}_{K'\geq K}$ is a martingale. To do so, we have to prove that for all $K’\geq K$, $X_{K’}=\sum_{k=K}^{K'} \alpha_k e_k$ satisfies the following two conditions:
	
	(i)	$\mathbb{E}[X_{K’+1}|X_{K’}] = X_{K’} $

	(ii)	$\mathbb{E}[\|X_{K’}\|^2]<\infty$.
	
	We know that $\mathbb{E}[e_k]=\mathbb{E}[g_k-\mathbb{E}[g_k|\mathcal{H}_k]]
	=\mathbb{E}_{\mathcal{H}_k}\Big[\mathbb{E}\Big[g_k-\mathbb{E}[g_k|\mathcal{H}_k]\Big|\mathcal{H}_k\Big]\Big]
	=0$
	by the law of total expectation. Hence, 
	$\mathbb{E}[ X_{K’+1}|X_{K’}]=\mathbb{E}\Big[\alpha_{K’+1} e_{K’+1}+ \sum_{k=K}^{K'} \alpha_k e_k \Big|\sum_{k=K}^{K'} \alpha_k e_k\Big]= 0+\sum_{k=K}^{K'} \alpha_k e_k=X_{K’}.$

	In addition, $e_k$ and $e_{k'}$ are uncorrelated for any $k\neq k'$ since (assuming $k>k'$)
	$\mathbb{E}\big[e_k^T e_{k’}\big]=\mathbb{E}\big[\mathbb{E}[e_k^T e_{k’}|\mathcal{H}_k]\big]= \mathbb{E}\big[e_{k’}\mathbb{E}[e_k^T |\mathcal{H}_k]\big]=0$. 
	
	Thus, 
	\begin{equation}\label{error_norm}
		\begin{split}
			\mathbb{E}(\|\sum_{k=K}^{K'}\alpha_k e_k\|^2)
			&=\mathbb{E}(\sum_{k=K}^{K'}\sum_{k'=K}^{K'}\alpha_k \alpha_{k'}\langle e_k, e_{k'}\rangle )\\
			&\overset{(a)}{=}\mathbb{E}(\sum_{k=K}^{K'}\|\alpha_k e_k\|^2)\\
			&\leq\sum_{k=K}^{\infty}\mathbb{E}(\alpha_k^2\|g_k-\mathbb{E}[g_k|\mathcal{H}_k]\|^2)\\
			&=\sum_{k=K}^{\infty}\alpha_k^2\mathbb{E}(\|g_k\|^2)-\mathbb{E}_{\mathcal{H}_k}(\|\mathbb{E}[g_k|\mathcal{H}_k]\|^2)\\
			&\leq\sum_{k=K}^{\infty}\alpha_k^2\mathbb{E}(\|g_k\|^2)\overset{(b)}{\leq}c_2\sum_{k=K}^{\infty}\alpha_k^2\overset{(c)}{<}\infty,\\
		\end{split}
	\end{equation}
	where $(a)$ is due to the uncorrelatedness $\mathbb{E}[\langle e_k,e_{k'}\rangle]= 0$, $(b)$ is by Lemma \ref{norm}, and $(c)$ is by Assumption \ref{step_sizes_1}.
	Therefore, both (i) and (ii) are satisfied and we can say that $\{\sum_{k=K}^{K'} \alpha_k e_k\}_{K'\geq K}$ is a martingale. This permits us to use Doob's martingale inequality \cite{doob}:
	
	For any constant $\nu>0$,
	\begin{equation}\label{doobm}
		\begin{split}
			\mathbb{P}(\sup_{K'\geq K}\|\sum_{k=K}^{K'}\alpha_k e_k\|\geq\nu) &\leq \frac{1}{\nu^2}\mathbb{E}(\|\sum_{k=K}^{K'}\alpha_k e_k\|^2)\\
			&\overset{(a)}{\leq}\frac{c_2}{\nu^2}\sum_{k=K}^{\infty}\alpha_k^2,
		\end{split}
	\end{equation}
	where $(a)$ is following the exact same steps as (\ref{error_norm}).
	
	Since $c_2$ is a bounded constant and $\lim_{K\rightarrow\infty}\sum_{k=K}^{\infty} \alpha_k^2= 0$ by Assumption \ref{step_sizes_1}, we get $\lim_{K\rightarrow\infty}\frac{c_2}{\nu^2}\sum_{k=K}^{\infty}\alpha_k^2 =0$ for any bounded constant $\nu$. Hence, the probability that $\|\sum_{k=K}^{K'}\alpha_k e_k\|\geq \nu$ also vanishes as $K\rightarrow\infty$, which concludes the proof.

	\subsection{Proof of Theorem \ref{th-ncvx}: Convergence Analysis} \label{th-ncvx-proof}
	By the $L$-smoothness assumption, the algorithm update step $\theta_{k+1}=\theta_k-\alpha_{k}g_k$, and knowing $g_k=e_k+\mathbb{E}[g_k|\mathcal{H}_k]$, we have
	\begin{equation*}
		\begin{split}
			F(\theta_{k+1})
			\leq &F(\theta_k)-\alpha_k\langle\nabla F(\theta_k), g_k\rangle +\frac{\alpha_k^2 L}{2}\|g_k\|^2\\
			= &F(\theta_k)-\alpha_k\langle\nabla F(\theta_k), e_k\rangle-c_1 \alpha_k\gamma_k\|\nabla F(\theta_k)\|^2\\&-c_1 \alpha_k\gamma_k\langle\nabla F(\theta_k), b_k\rangle+\frac{\alpha_k^2 L}{2}\|g_k\|^2\\
			\overset{(a)}{\leq} &F(\theta_k)-\alpha_k\langle\nabla F(\theta_k), e_k\rangle-c_1 \alpha_k\gamma_k\|\nabla F(\theta_k)\|^2\\+&\frac{c_1 \alpha_k\gamma_k}{2}\|\nabla F(\theta_k)\|^2+\frac{c_1 \alpha_k\gamma_k}{2}\|b_k\|^2  +\frac{\alpha_k^2 L}{2}\|g_k\|^2\\
		\end{split}
	\end{equation*}
	\begin{equation}\label{cvanalysis}
		\begin{split}
			\overset{(b)}{\leq} &F(\theta_k)-\alpha_k\langle\nabla F(\theta_k), e_k\rangle-\frac{c_1 \alpha_k\gamma_k}{2}\|\nabla F(\theta_k)\|^2 \\&+\frac{c_1 c_3^2 \alpha_k\gamma_k^3}{2}  +\frac{\alpha_k^2 L}{2}\|g_k\|^2
		\end{split}
	\end{equation}
	where $(a)$ is by $-\langle a,b\rangle\leq \frac{1}{2}\|a\|^2+\frac{1}{2}\|b\|^2$ and $(b)$ is by Lemma \ref{bias-norm-lemma}.
	
	By taking the telescoping sum, we get 
	\begin{equation*}
		\begin{split}
			&F(\theta^*) \leq F(\theta_{K+1})
			\leq F(\theta_0)-\frac{c_1}{2}\sum_{k=0}^{K}\alpha_k\gamma_k\|\nabla F(\theta_k)\|^2 \\&-\sum_{k=0}^{K}\alpha_k\langle\nabla F(\theta_k), e_k\rangle+\frac{c_1 c_3^2 }{2}\sum_{k=0}^{K}\alpha_k\gamma_k^3+\frac{c_2L}{2}\sum_{k=0}^{K}\alpha_k^2\|g_k\|^2\\
		\end{split}
	\end{equation*}
	Hence, knowing that $\delta_0=F(\theta_0)-F(\theta^*)$,
	\begin{equation}\label{ineq_sum}
		\begin{split}
			\sum_{k=0}^{K}\alpha_k\gamma_k\|\nabla F(\theta_k)\|^2	
			\leq &\frac{2\delta_0}{c_1}-\frac{2}{c_1}\sum_{k=0}^{K}\alpha_k\langle\nabla F(\theta_k), e_k\rangle \\
			&+c_3^2\sum_{k=0}^{K}\alpha_k\gamma_k^3+\frac{ c_2 L}{c_1}\sum_{k=0}^{K}\alpha_k^2\|g_k\|^2\\
		\end{split}
	\end{equation}
	By Assumption \ref{local_fcts}, $\|\nabla F(\theta_k)\|$ is bounded for any $\theta_k\in\mathbb{R}^d$ and by taking the summation in (\ref{doobm}) between $0$ and $\infty$, we have $\lim_{K\rightarrow\infty}\|\sum_{k=0}^{K}\alpha_k\langle\nabla F(\theta_k),e_k\rangle\|<\infty$. By Assumption \ref{step_sizes_1},
 	$\lim_{K\rightarrow\infty}\sum_{k=0}^{K}\alpha_k\gamma_k^3<\infty$.
	
	To prove the finiteness of $\sum_{k=0}^{K}\alpha_k^2\|g_k\|^2$, we let $X_k$ be a centered Gaussian process. We know that for any $\sigma^2$-subgaussian random variables $X_1, \ldots, X_K$, we have 
	
	$$\mathbb{P}\left[\sup_{1\leq k\leq K} X_k\geq \sqrt{2\sigma^2 (\log K +t)}\right]\leq e^{-t}$$
	since let $u := \sqrt{2 \sigma^2(\log K+t)}$,
	\begin{equation*}
		\begin{split}
			\mathbb{P}\left[\sup_{1\leq k\leq K} X_k\geq u\right] &= \mathbb{P}\left[\exists k, X_k\geq u\right] \\ &\leq \sum_{k=1}^{K}\mathbb{P}\left[ X_k\geq u\right]\leq K e^{-\frac{u^2}{2\sigma^2}} = e^{-t}.
		\end{split}
	\end{equation*}
	
	Then, for $t=c \log K$ with $c>1$, 
	$\mathbb{P}\left[\sup_{1\leq k\leq K} X_k\geq \sqrt{2\sigma^2(1+c)\log K }\right]\leq \frac{1}{K^c}$
	and 
	$$\sum_{K=1}^{\infty}\mathbb{P}\left[\sup_{1\leq k\leq K} X_k\geq \sqrt{2\sigma^2(1+c) \log K }\right]\leq \sum_{K=1}^{\infty}\frac{1}{K^c}<\infty.$$
	
	By Borel-Cantelli Lemma, we have $$\mathbb{P}\left(\lim_{K\rightarrow\infty}\sup \{\sup_{1\leq k\leq K} X_k\geq \sqrt{2\sigma^2(1+c)\log K }\}\right)=0.$$
	
	Then, w.p.1 $\exists K'<\infty$ s.t. $\forall K\geq K'$, $\sup_{1\leq k\leq K} X_k\leq \sqrt{2\sigma^2(1+c) \log K }$ with $c>1$.
	
	Thus, the supremum of the Gaussian elements of $\|g_k\|$, i.e., $h_{i,k+1}, h_{i,k}$, $n_{i,k}$, and $n_{i,k+1}$, at worst grow as $O(\sqrt{log (k+1)})$, and hence the upper bound on $\|g_k\|$ grows as $c_4 \sqrt{\log(k+1)}$, where $c_4 = \beta_2 N\big(\frac{C}{\sigma_h}+\sigma_n\big) \sqrt{2(1+c)}$:
	
	\begin{equation*}
		\begin{split}
			\|g_k\|
			&=\Big\|\Phi_k \Big(\sum_{i=1}^{N}h_{i,k+1}\frac{f_i\big(\theta_k', \xi_{k+1}\big)}{\sigma_h^2}+n_{i,k+1}\Big)\Big\|\\
			&\leq \|\Phi_k \|\Big(\sum_{i=1}^{N} \Big\|h_{i,k+1}\frac{C}{\sigma_h^2}\Big\|+\|n_{i,k+1}\|\Big)\\
			&\leq \beta_2 N \bigg(\frac{C}{\sigma_h^2}\sqrt{2\sigma_h^2(1+c)\log(k+1)}\\&\hspace{2cm}+\sqrt{2\sigma_n^2(1+c)\log(k+1)}\bigg)\\
			&= \beta_2 N\big(\frac{C}{\sigma_h}+\sigma_n\big) \sqrt{2(1+c)\log(k+1)}
		\end{split}
	\end{equation*}
	
	Then, we write $\sum_{k=0}^{K}\alpha_k^2\|g_k\|^2 = \sum_{k=0}^{K'}\alpha_k^2\|g_k\|^2+ \sum_{k=K'}^{K}\alpha_k^2\|g_k\|^2$, where $\sum_{k=0}^{K'}\alpha_k^2\|g_k\|^2 <\infty$ for $K'< \infty$.
	
	We know that $\forall \epsilon>0$, $\log(k+1)\leq \frac{1}{\epsilon}(k+1)^{\epsilon}$. Thus, by Assumption \ref{step_sizes_1}, for $2\upsilon_1=1+\epsilon'$,
	\begin{equation}
		\begin{split}
			&\lim_{K\rightarrow\infty}\sum_{k=K'}^{K}\alpha_k^2\|g_k\|^2 \\ &\leq \lim_{K\rightarrow\infty}c_4^2\sum_{k=K'}^{K}\alpha_k^2 \log(k+1) \\ &\leq \lim_{K\rightarrow\infty}c_4^2\sum_{k=K'}^{K}\frac{1}{(k+1)^{1+\epsilon'}}\times \frac{1}{\epsilon}(k+1)^{\epsilon}  <\infty, \\ &\hspace{5cm}\;\;\forall \epsilon'>\epsilon>0.
		\end{split}
	\end{equation}	 
	We conclude that 
	\begin{equation}\label{nablaF_1}
		\lim_{K\rightarrow\infty}\sum_{k=0}^{K}\alpha_k\gamma_k\|\nabla F(\theta_k)\|^2	<\infty.
	\end{equation}
	Moreover, since the series $\sum_k\alpha_k\gamma_k$ diverges by Assumption \ref{step_sizes_1}, we have $\lim_{k\rightarrow\infty}\inf \|\nabla F(\theta_k)\|=0.$
	
	To prove that $\lim_{k\rightarrow\infty}\|\nabla F(\theta_k)\|=0$, we consider the hypothesis H) $\lim_{k\rightarrow\infty}\sup \|\nabla F(\theta_k)\|\geq \rho$ for an arbitrary $\rho>0$.
	
	Assume (H) to be true. Then, we can always find an arbitrary subsequence $\big(\|\nabla F(\theta_{k_l})\|\big)_{l\in\mathbb{N}}$ of $\|\nabla F(\theta_k)\|$, such that $\|\nabla F(\theta_{k_l})\|\geq \rho - \varepsilon$, $\forall l$, for $\rho - \varepsilon>0$ and $\varepsilon>0$.
	
	Then, by the $L$-smoothness property and applying the descent step of the algorithm, 
	\begin{equation}
		\begin{split}
			\|\nabla F(\theta_{k_l+1})\|&\geq \|\nabla F(\theta_{k_l})\|-\|\nabla F(\theta_{k_l+1})-\nabla F(\theta_{k_l})\| \\
			&\geq \rho - \varepsilon-L\|\theta_{k_l+1}-\theta_{k_l}\|\\
			&= \rho - \varepsilon-L\alpha_{k_l}\|g_{k_l}\|\\
			&\geq \rho - \varepsilon-Lc_4\sqrt{\log(k_l+1)}\alpha_{k_l},
		\end{split}
	\end{equation}

	Since $k_l\rightarrow\infty$ as $l\rightarrow\infty$, we can always find a subsequence of $(k_{l_p})_{p\in\mathbb{N}}$ such that $k_{l_{p+1}}-k_{l_p}>1$. As $\alpha_{k_l}$ is vanishing, we consider $(k_l)_{l\in\mathbb{N}}$ starting from $\alpha_{k_l}<\frac{\rho - \varepsilon}{Lc_4\sqrt{\log(k_l+1)}}$. As a reminder, we can always find $\epsilon>0$ such that $\log(x)\leq \frac{1}{\epsilon}x^{\epsilon}$. Thus,
	\begin{equation}
		\begin{split}
			&\sum_{k=0}^{\infty} \alpha_{k+1}\gamma_{k+1}\|\nabla  F(\theta_{k+1})\|^2 \\ \geq &(\rho - \varepsilon)^2\sum_{k=0}^{\infty}\alpha_{k+1}\gamma_{k+1}\\&-2(\rho - \varepsilon)L c_4\sum_{k=0}^{\infty}\alpha_{k+1}\gamma_{k+1}\alpha_{k}\sqrt{\log(k+1)}\\&+ L^2c_4^2\sum_{k=0}^{\infty}\alpha_{k+1}\gamma_{k+1}\alpha_{k}^2\log(k+1)\\
			>&+\infty.
		\end{split}
	\end{equation}
	as the first series diverges, and the second and the third converge by Assumption \ref{step_sizes_1}.
	This implies that the series $\sum_{k} \alpha_{k}\gamma_{k}\|\nabla  F(\theta_{k})\|^2$ diverges. This is a contradiction as this series converges almost surely by (\ref{nablaF_1}). Therefore, hypothesis (H) cannot be true and $\|\nabla F(\theta_{k})\|$ converges to zero almost surely.
	
	\subsection{Proof of Theorem \ref{th-ncvx-rate}: Convergence Rate} \label{th-ncvx-rate-proof}
	Taking the expectation on both sides of (\ref{ineq_sum}) results in $\mathbb{E}[\langle\nabla F(\theta_k), e_k\rangle] = \mathbb{E}_{\mathcal{H}_k}\Big[\mathbb{E}\big[\langle\nabla F(\theta_k), e_k\rangle\big|\mathcal{H}_k\big]\Big]=0$. Then, substituting by the upper bound of Lemma \ref{norm},
	
	\begin{equation}\label{nablaF_}
		\begin{split}
			\sum_{k=0}^{K}\alpha_k\gamma_k\mathbb{E}[\|\nabla F(\theta_k)\|^2]	
			&\leq \frac{2}{c_1}\delta_0 +c_3\sum_{k=0}^{K}\alpha_k\gamma_k^3+\frac{ L c_2}{c_1}\sum_{k=0}^{K}\alpha_k^2\\
		\end{split}
	\end{equation}
	
	Let $\alpha_k$ and $\gamma_k$ have the forms given in Example \ref{eg}.	We know that, $\forall K>0$, 
	$\sum_{k=0}^{K}\alpha_k\gamma_k^3 = \alpha_0 \gamma_0^3 +\sum_{k=1}^{K}\alpha_k\gamma_k^3\leq \alpha_0 \gamma_0^3\Big(1 + \int_{0}^{K}(x+1)^{-\upsilon_1-3\upsilon_2}dx\Big)
			\leq \alpha_0 \gamma_0^3\Big(1 + \frac{1}{\upsilon_1+3\upsilon_2-1}\Big)
			= \alpha_0 \gamma_0^3\Big(\frac{\upsilon_1+3\upsilon_2}{\upsilon_1+3\upsilon_2-1}\Big)$.
			
	Similarly, $\sum_{k=0}^{K}\alpha_k^2 \leq \alpha_0^2 \big(\frac{2\upsilon_1}{2\upsilon_1-1}\big)$.
	Next, when further $0<\upsilon_1+\upsilon_2<1$,
	$ \sum_{k=0}^{K}\alpha_k\gamma_k \geq \alpha_0\gamma_0\int_{0}^{K+1} (x+1)^{-\upsilon_1-\upsilon_2} dx= \frac{\alpha_0\gamma_0}{(1-\upsilon_1-\upsilon_2)} \Big((K+2)^{1-\upsilon_1-\upsilon_2}-1\Big)$.
		
	Thus, making use of inequality (\ref{nablaF_}),
	\begin{equation}\label{sth}
		\frac{\sum_{k}\alpha_k\gamma_k\mathbb{E}[\|\nabla F(\theta_k)\|^2]}{\sum_k \alpha_k\gamma_k}
		\leq \frac{A(1-\upsilon_1-\upsilon_2)}{(K+2)^{1-\upsilon_1-\upsilon_2}-1},
	\end{equation}
	with $A= \frac{2}{c_1\alpha_0\gamma_0}\delta_0+ c_3^2\gamma_0^2\big(\frac{\upsilon_1+3\upsilon_2}{\upsilon_1+3\upsilon_2-1}\big)+\frac{ L c_2\alpha_0}{c_1\gamma_0}  \big(\frac{2\upsilon_1}{2\upsilon_1-1}\big)$.
	
	We see that $\upsilon_1=\frac12$ and $\upsilon_2=\frac16$ are the optimal choice for the time-varying part of the rate, i.e., for $O(\frac{1}{\sqrt[3]{K}})$. However, to prevent the constant part from growing too large, we choose slightly larger exponents of $\upsilon_1=\frac12+\frac{\epsilon}{2}$ and $\upsilon_2=\frac16+\frac{\epsilon}{2}$, where $\epsilon$ is a small strictly positive value. This will result in a rate of $O\big(\frac{1}{K^{\frac13-\epsilon}}\big)$.
	
	\section{Performance of 1P-ZOFL vs SNR}\label{experiments}	
	An important remark is that high SNR is generally needed when we must decode the information in the received signal. In our case, nothing is decoded; there is no channel estimation or gradient extraction from the received signal. Rather, the received signal is fed directly into the learning (the channel is part of the learning). Ultimately, the amount of noise present in the system does not affect our algorithms' convergence: Examining (\ref{grdt_norm_1}) and (\ref{bias_norm}), we see that the noise variance $\sigma_n^2$ might only increase the norms of the estimate and bias, but if we refer to (\ref{cvanalysis}), we find that both terms are multiplied by step sizes, and we can counter the noise effect by decreasing the step sizes' constant parts (i.e, in the terms $\frac{c_1 \alpha_k\gamma_k}{2}\|b_k\|^2$ and $\frac{\alpha_k^2 L}{2}\|g_k^{(1P)}\|^2$). For the different plots in Fig. \ref{iid2}, we decrease $\alpha_0$ and $\gamma_0$ when we increase $\sigma_n^2$. However, this change of step sizes generally affects the rate of convergence as they enter in the structure of $g_k$ and $b_k$ themselves and as can be seen in (\ref{sth}), smaller $\alpha_0$ and $\gamma_0$ increases the first term in the constant part of the upper bound on the convergence rate. Thus, the effect shown on the plots in Fig. \ref{iid2} is logical given the changes of $\alpha_0$ and $\gamma_0$.



	

\end{document}